\documentclass[11pt,a4paper]{article}
\usepackage[dvipsnames]{xcolor}
\usepackage[hyperref]{acl2020}

\usepackage{times, tablefootnote}
\usepackage{latexsym}

\aclfinalcopy 
\usepackage{graphicx}
\usepackage{soul, amsmath, amssymb, multirow}
\usepackage[inline]{enumitem}
\usepackage{lipsum}
\usepackage{array}
\usepackage{tabu}
\usepackage{booktabs}
\usepackage{float}
\usepackage{subcaption}
\restylefloat{table}

\title{Improving Disfluency Detection by Self-Training a Self-Attentive Model}

\author{Paria Jamshid Lou$^1$ and Mark Johnson$^{2, 1}$\\
  $^1$Department of Computing, Macquarie University \\
  $^2$Oracle Digital Assistant, Oracle Corporation\\
  \texttt{$^1$paria.jamshid-lou@hdr.mq.edu.au}\\
  \texttt{$^2$mark.mj.johnson@oracle.com} \\}
\date{}

\definecolor{editcolor}{rgb}{0.10, 0.56, 0.6}
\definecolor{intcolor}{rgb}{0.65, 0.74, 0.85}

\begin{document}
\maketitle
\begin{abstract}
Self-attentive neural syntactic parsers using contextualized word embeddings (e.g. ELMo or BERT) currently produce state-of-the-art results in joint parsing and disfluency detection in speech transcripts. Since the contextualized word embeddings are pre-trained on a large amount of unlabeled data, using additional unlabeled data to train a neural model might seem redundant. However, we show that self-training --- a semi-supervised technique for incorporating unlabeled data --- sets a new state-of-the-art for the self-attentive parser on disfluency detection, demonstrating that self-training provides benefits orthogonal to the pre-trained contextualized word representations. We also show that ensembling self-trained parsers provides further gains for disfluency detection. 
\end{abstract}



 
\section{Introduction}
Speech introduces challenges that do not appear in written text, such as the presence of disfluencies. Disfluency refers to any interruptions in the normal flow of speech, including false starts, corrections, repetitions and filled pauses.~\citet{shri:94} defines three distinct parts of a speech disfluency, referred to as the \textit{reparandum}, the \textit{interregnum} and the \textit{repair}. As illustrated in the example below, the reparandum \textit{The first kind of invasion of} is the part of the utterance that is replaced or repaired, the interregnum \textit{uh I mean} (which consists of a filled pause \textit{uh} and a discourse marker \textit{I mean}) is an optional part of the disfluency, and the repair \textit{the first type of privacy} replaces the reparandum. The fluent version is obtained by removing the reparandum and the interregnum. 
                                                                                                                                                                                                                                                                                                                                                                                      
\begin{equation*}  \label{ex:1}
\centering 
\begin{array}{l}
\mbox{\it ~~~~~~~}\hspace{-0.05cm}\overbrace{\mbox{\it\strut \textcolor{editcolor}{The first kind of invasion of }}}^{\mbox{\scriptsize \textcolor{editcolor}{reparandum}}}\hspace{-0.13cm}\strut\overbrace{\mbox{\it \textcolor{intcolor}{uh I mean}\strut}}^{\mbox{\scriptsize \textcolor{intcolor}{interregnum}}}\\ \underbrace{\mbox{\it\strut the first type of privacy}}_{\mbox{\scriptsize repair}} \mbox{\it seemed invaded to me}
\end{array}
\end{equation*}

\begin{figure} [h]
\centering 
\includegraphics[width=0.48\textwidth]{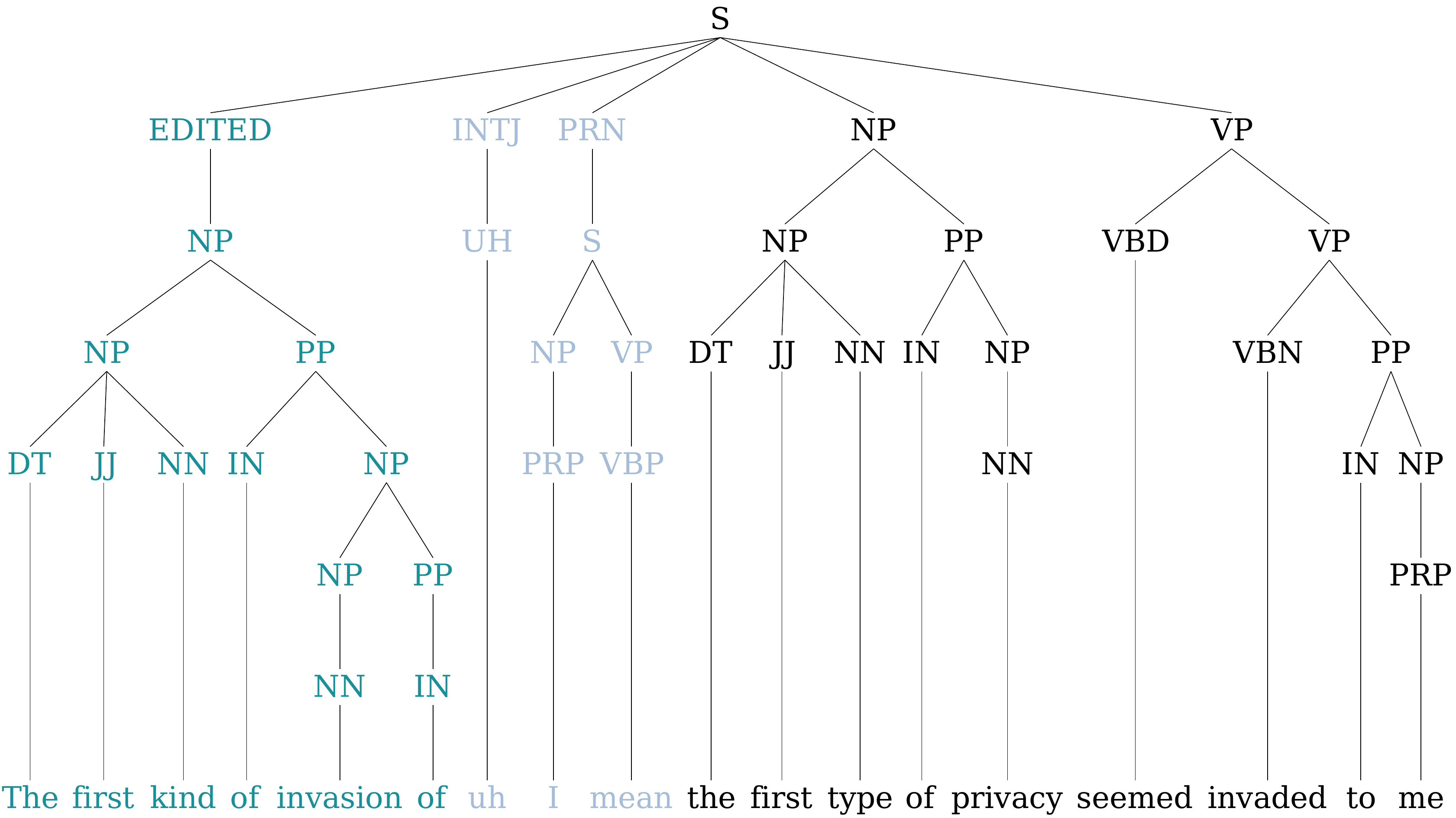}\caption{A parse tree from the Switchboard corpus, where reparandum \emph{The first kind of invasion of}, filled pause \emph{uh} and discourse marker \emph{I mean} are dominated by EDITED, INTJ and PRN nodes.}
\label{fig:01}
\end{figure}

This paper will focus on joint disfluency detection and constituency parsing of transcribed speech. In the Switchboard treebank corpus~\citep{godfrey:93, mit:99}, which is a standard corpus for parsing studies on conversational speech, the \textit{reparanda}, \textit{filled pauses} and \textit{discourse markers} are dominated by EDITED, INTJ and PRN nodes, respectively (see Figure~\ref{fig:01}). Filled pauses and discourse markers belong to a finite set of words and phrases, so INTJ and PRN nodes are trivial to detect~\citep{john:04}. Detecting EDITED nodes, however, is challenging and is the main focus of disfluency detection models.

\citet{jam:19} showed that a self-attentive constituency parser achieves state-of-the-art results for joint parsing and disfluency detection. They observed that because the Switchboard trees include both syntactic constituency nodes and EDITED nodes that indicate disfluency, training a parser to predict the Switchboard trees can be regarded as multi-task learning (where the tasks are syntactic parsing and identifying disfluencies). In this paper, we extend the multi-task learning in~\citet{jam:19} to explore the impact of self-training~\citep{mc:06} and ensembling~\citep{kita:19} on the performance of the self-attentive parser. We aim to answer two questions about the state-of-the-art self-attentive parser:

\begin{itemize}
\item \emph{Does self-training improve the performance of the self-attentive parser on disfluency detection?} Self-training is a semi-supervised technique for incorporating unlabeled data into a new model, where an existing model trained on manually labeled (i.e. gold) data is used to label unlabeled data. The automatically (i.e. silver) labeled data are treated as truth and combined with the gold labeled data to re-train a new model~\citep{mc:06, cho:16}. Since neural models use rich representations of language pre-trained on a large amount of unlabeled data~\citep{peters:18, dev:19}, we might expect that self-training adds no new information to the self-attentive parser. Surprisingly, however, we find that \textbf{self-training improves disfluency detection f-score of the BERT-based self-attentive parser}, demonstrating that self-training provides benefits orthogonal to the pre-trained contextualized embeddings.

\item \emph{Does ensembling improve disfluency detection in speech transcripts?} Ensembling is a commonly used technique for improving parsing where scores of multiple instances of the same model trained on the same or different data are combined at inference time~\citep{dyer:16, fried:17, kita:19}. We expect ensembling parsers to improve the performance of the model on disfluency detection, too. We show \textbf{ensembling four self-trained parsers (using different BERT word representations) via averaging their span label scores increases disfluency detection f-score} in comparison with a single self-trained parser. 

\end{itemize}



\section{Related Work}
Parsing speech transcripts is challenging for conventional syntactic parsers, mainly due to the presence of disfluencies. In disfluent sentences, the relation between reparandum and repair is different from other words in the sentence. The repair is usually a ``rough copy'' of the reparandum, using the same or similar words in roughly the same word order\footnote{For example in Figure~\ref{fig:01}, the reparandum \emph{The first kind of invasion of} and the repair \emph{the first type of privacy} are ``rough copies'' of each other.}~\citep{char:01}. Designed to capture tree-like structures, conventional syntactic parsers fail to detect ``rough copies'' which are strong indicators of disfluency. Moreover, the reparandum and repair often do not form a syntactic phrase, which makes detecting the reparandum even harder. For these reasons, specialized disfluency detection models were developed to remove disfluencies prior to parsing~\citep{char:01, kahn:05, lease:06} or special mechanisms were added to parsers to handle disfluencies~\citep{ras:13, hon:14, yoshi:16}. Conventional parsing based models can use the syntactic location of the disfluency as a feature in a reranker~\citep{john:04a}. A similar gain can be achieved in neural models by training a joint parsing and disfluency detection model. In this multi-task learning setting, syntactic information helps the neural model detect disfluencies more accurately~\citep{jam:19}.  

State-of-the-art results for disfluency detection have been reported for Transformer models using contextualized embeddings (e.g. ELMo and BERT)~\citep{jam:19, tran:19, dong:19}. The self-attention mechanism of the Transformer is apparently effective for capturing ``rough copy'' dependencies between words. A recent study shows that prosody slightly improves the parsing performance of the self-attentive model over the text-only model, especially in long sentences~\citep{tran:19}. In this paper, we use a self-attentive model for joint disfluency detection and constituency parsing.

Disfluency detection models are usually trained and evaluated on the Switchboard corpus. Switchboard is the largest disfluency annotated dataset. However, only 5.9\% of the words in the Switchboard are disfluent~\citep{char:01}. To mitigate the scarcity of labeled data, some studies have leveraged additional data by using: \begin{enumerate*}[label=(\roman*)] \item contextualized embeddings pre-trained on enormous amount of unlabeled data~\citep{jam:19, tran:19, bach:19} and \item synthetic data generated by adding noise in the form of disfluencies to fluent sentences (e.g. repeating, deleting or inserting words in a sentence)~\citep{wang:18, bach:19, dong:19}.\end{enumerate*} By contrast, this paper focuses on self-training, which is a simple semi-supervised technique that has been effective in different NLP tasks, including parsing~\citep{mc:06, clark:18, drog:18}. To our best knowledge, this is the first work that investigates self-training a neural disfluency detection model.

Another technique commonly used for improving parsing is ensembling. Ensembling is a model combination method, where scores of multiple models (they can be the same or different models, trained on the same or different data, with different random initializations) are combined in some way~\citep{dyer:16, cho:16, fried:17}. The state-of-the-art for parsing written text is an ensemble of four BERT-based self-attentive parsers, where the parsers are combined by averaging their span label scores~\citep{kita:19}. While ensembling is widely used in parsing, it has not been investigated for disfluency detection. In this paper, we also explore the impact of ensembling several parsing based disfluency detection models on disfluency detection performance.



\section{Model} \label{sec:mo}
Following~\citet{jam:19}, we use a self-attentive constituency parser for joint disfluency detection and syntactic parsing\footnote{The code is available at: \url{https://github.com/pariajm/joint-disfluency-detector-and-parser}}. The parsing model is based on the architecture introduced by~\citet{kita:18}, which is state-of-the-art for 
\begin{enumerate*}[label=(\roman*)] \item parsing written texts~\citep{kita:19, fri:19}, \item parsing transcribed speech~\citep{tran:19}, and \item joint parsing and disfluency detection~\citep{jam:19}\end{enumerate*}. 

The self-attentive parser assigns a score $s(T)$ to each tree $T$ by calculating the sum of the potentials on its labeled constituent spans:

\begin{equation} \label{eq:02}
s(T)=\displaystyle\sum_{(i,j,l) \in T} s(i,j,l)
\end{equation}
where $s(i,j,l)$ is the score of a constituent beginning at string position $i$ ending at position $j$ with label $l$. The input to the parser is a sequence of vectors corresponding to the sequence of words in a sentence followed by one or more self-attention layers. For each span $(i, j)$, a hidden vector $h_{ij}$ is constructed by subtracting the representations of the start and end of the span.  A span classifier, including two fully connected layers followed by a non-linearity, assigns labeling scores $s(i, j, .)$ to each span. Then, the highest scoring parse tree is found for a given sentence as follows:
\begin{eqnarray}
\hat{T} = \operatorname*{argmax}_{T} s(T)
\end{eqnarray}
using a modified CYK algorithm. The parser introduced in~\citet{kita:18} relies on an external POS tagger to predict preterminal labels, but because the parser's accuracy does not decrease when no external POS tagger is used, we use their parser here without an external POS tagger (hence, all the preterminal labels are UNK). For more details, see~\citet{kita:18}.

\subsection{Contextualized Embeddings}
We incorporate BERT~\citep{dev:19} in our self-attentive parser by fine-tuning the parameters as part of the training process. Following~\citet{kita:19}, we apply a learned projection matrix on the output of BERT to project the vectors to our desired dimensionality. The representations are then fed into the parser. BERT learns the representations for sub-word units, so to extract the word representations, we consider the representations of the last sub-word unit for each word in the sentence~\citep{kita:19}.

\subsection{Self-Training}
We train the self-attentive parser on the Penn Treebank-3 Switchboard corpus which contains gold disfluency labeled parse trees~\cite{godfrey:93, mit:99}. Using the trained model, we parse unlabeled data and add the silver parse trees to the gold Switchboard training data and re-train the self-attentive parser using the enlarged training set. The unlabeled data we use include Fisher Speech Transcripts Part 1~\citep{cieri:04} and Part 2~\citep{cieri:05}. Table~\ref{tab:11} summarizes the different datasets used to train the self-attentive parser.

\begin{table}[H]
\begin{center}
\begin{tabular}{lccc} \\ \midrule 
\bf Dataset &  \bf Labels & \bf{\# Sents} & \bf{\# Words}  \\  \midrule
SWB & gold & 98k & 733k  \\  
Fisher & silver & 835k &  14m  \\\midrule 
\end{tabular}
\end{center}
\caption{Summary of the datasets used to train the self-attentive parser.}\label{tab:11}
\end{table}




\section{Experiments}
Following Charniak and Johnson~\shortcite{char:01}, we split the Switchboard into training, dev and test sets as follows: training data consists of the sw[23]$\ast$.mrg files, dev data consists of the sw4[5-9]$\ast$.mrg files and test data consists of the sw4[0-1]$\ast$.mrg files. All partial words\footnote{Words tagged as ``XX'' or words ending in ``-''} and punctuations are removed from the data, as they are not available in realistic ASR applications~\citep{john:04}.

\subsection{Baseline}
Our baseline is the self-attentive parser trained on the gold Switchboard corpus with BERT word representations. The BERT-based~parser~is the current state-of-the-art, providing a very strong baseline for our work. We trained different versions of the baseline parser using four different BERT models, namely BERT$_{\text{BASE}~ \text{[cased}\mid\text{uncased]}}$ and BERT$_{\text{LARGE}~\text{[cased}\mid\text{uncased]}}$, and then selected the best model i.e. BERT$_{\text{BASE}~ \text{[cased]}}$ on the Switchboard dev set. We also tuned the hyperparameters by optimizing for performance on parsing EDITED nodes $F(S_E)$. Preliminary experiments on the Switchboard dev set showed that the hyperparameters given by~\citet{kita:19} perform well; therefore, this is what we used here. Since random seeds lead to different results, in this paper we report average scores across 5 runs of each model initialized with different random seeds.

\subsection{Evaluation Metrics}
We evaluate the self-attentive parser in terms of parsing accuracy, as well as disfluency detection. Since certain words are identified as EDITED in the parse tree, we can measure how well a parser classifies words as EDITED.  We can also evaluate how accurately the parser can identify all disfluency words, i.e., the words dominated by EDITED, INTJ or PRN nodes. Therefore, we report \emph{precision} (P), \emph{recall} (R) and \emph{f-score} (F) for both constituent spans (S) and word positions (W), where each word position is treated as labeled by all the constituents containing that word. We also report the result for subsets of constituent spans and word positions: \begin{enumerate*}[label=(\roman*)] \item S$_\text{E}$, the set of constituent spans labeled EDITED, \item W$_\text{E}$, the set of word positions dominated by one or more EDITED nodes, and \item W$_\text{EIP}$, the set of word positions dominated by one or more EDITED, INTJ or PRN nodes. For more details, see~\citet{jam:19}.\end{enumerate*}

\subsection{Varying Amount of Silver Training Data} \label{4.4.1}
\begin{figure}[h]
 \centering
\includegraphics[width=0.46\textwidth]{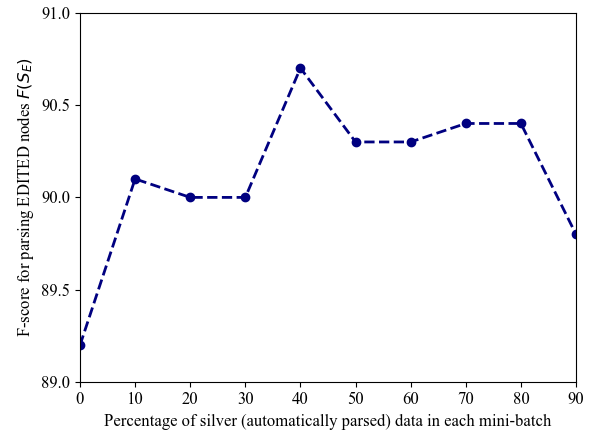}
\caption{EDITED node f-score $F(S_{E})$ of the BERT-based self-attentive parser as a function of percentages of training data in each mini-batch sourced from silver Fisher trees. }
\label{fig:02}
\end{figure}

To find the optimal proportion of additional silver training data, we select \emph{n} percent (ranging from $10\%$ to $90\%$) of the training data in each mini-batch from silver parse trees and the rest from the gold ones. This has the same effect as re-weighting the main gold corpus as in~\citet{mc:06}. The results for using different proportions of the silver parse trees are presented in Figure~\ref{fig:02}.  The BERT-based parser self-trained with 40\% silver Fisher trees and 60\% gold Switchboard trees is our best model. In other words, for a batch size of 30, in each mini-batch 12 parse trees come from the silver Fisher data and 18 parse trees from the gold Switchboard. All self-training results in this paper use this proportion of gold and silver parse trees. 

\subsection{Does self-training improve the performance of the self-attentive parser?}\label{3.4}
Tables~\ref{tab:01} and~\ref{tab:02} compare the baseline and the self-trained parser in terms of parsing and disfluency detection. The parser self-trained on the silver Fisher data increases parsing and disfluency detection performance, indicating the BERT-based model benefits from additional silver labeled data. Self-training is especially effective for recognizing EDITED disfluency nodes ($1.5\%$ increase in f-score). Only $5.9\%$ of the words in the Switchboard are disfluent, and BERT is only trained on fluent texts such as books and Wikipedia, so the baseline parser may be starved of disfluent training examples. As a result, self-training on a corpus of conversational speech may compensate for the scarcity of disfluent gold data. To explore this, we tried self-training on a wide variety of fluent clean datasets, including Gigaword 5 (which is an unlabelled newswire corpus) and WSJ and Brown (which include gold parse trees of written text), but the performance did not improve significantly. This suggests that the parser benefits more from additional in-domain (i.e. conversational) silver data than additional out-of-domain (i.e. written) silver/gold data. Moreover, if we learn the embeddings as part of training instead of using pre-trained BERT, EDITED word f-score would drop from 90.9\% to 86.4\% and self-training on Fisher leads to little improvement (0.2\% increase in EDITED word f-score compared to 1.5\% improvement when using BERT). This suggests that self-training works well when the baseline model is powerful enough to predict accurate silver labels. 


\begin{table}[h]
\begin{center}
\begin{tabular}{lccc} \\ \midrule 
\bf Parsing &  \bf F(S$_\text{E}$) &  \bf F(S$_\text{EIP}$)  & \bf F(S)  \\  \midrule
Baseline & $89.2$ & $95.6$ &  $93.5$   \\  
Self-trained & $90.7$ & $96.2$ &  $93.9$  \\
 \midrule
\end{tabular}
\end{center}
\caption{Parse f-score for EDITED node \emph{F($\text{S}_\text{E}$)}, for EDITED, INTJ and PRN nodes \emph{F($\text{S}_\text{EIP}$)} and for all constituent spans \emph{F(S)} on the Switchboard dev set for the baseline parser and the parser trained on the silver Fisher data.}\label{tab:01}
\end{table}

\begin{table}[h]
\begin{center}
\begin{tabular}{p{2.8cm}cc} \\
\midrule \bf Disfluency &  \bf F(W$_\text{E}$) &  \bf F(W$_\text{EIP}$)  \\ \midrule
Baseline & $90.9$ & $95.3$  \\ 
Self-trained & $92.4$ &  $96.0$   \\ \midrule
\end{tabular}
\end{center}
\caption{EDITED word f-score \emph{F(W$_{E}$)}, EDITED, INTJ and PRN word f-score \emph{F(W$_{EIP}$)} on the Switchboard dev set for the baseline parser and the parser trained on the silver Fisher data.}\label{tab:02}
\end{table}

\vspace{0.2cm}
To further investigate the influence of self-training on disfluency detection, we randomly select $100$ sentences containing disfluencies from the Switchboard dev set. We categorize disfluencies into~\emph{repetition}, \emph{correction} and \emph{restart} according to Shriberg's~\citeyearpar{shri:94} typology of speech repairs. \emph{Repetitions} are repairs where the reparandum and repair portions of the disfluency are identical, while \emph{corrections} are where the reparandum and repairs differ (which are much harder to detect). \emph{Restarts} are where the speaker abandons a sentence and starts a new one (i.e. the repair is empty). As Table~\ref{tab:13} shows, the self-trained parser outperforms the baseline in detecting all types of disfluency. It especially has a better performance at detecting corrections and restarts which are more challenging types of disfluency in comparison with repetitions. 

\begin{table}[H]
	\begin{center}
		\begin{tabular}{lcccc}
			\midrule \bf Model & \bf  Rep. & \bf Cor. & \bf  Res. & \bf All \\ \midrule 
			Baseline &  $97.0$ & $80.6$ &  $82.0$ & $89.2$ \\ 
	Self-trained & $97.3$ & $88.6$ &  $87.8$ & $92.9$ \\
			\midrule
		\end{tabular}
	\end{center}
	\caption{EDITED word f-score $F(W_\text{E})$ for different types of disfluency on a subset of the Switchboard dev set containing $158$ disfluent structures --- including $90$ repetitions (Rep.), $54$ corrections (Cor.) and $14$ restarts (Res.). }\label{tab:13} 
\end{table}

\begin{table*}[t]
\begin{center}
\begin{tabular}{p{0.1cm}p{2cm}p{12.86cm}}
 \midrule \bf \# & \bf Model &  \bf{EDITED Disfluency Labels}\\ \midrule

\multirow{3}{*}{\bf 1} & \bf{Gold} & \small{{\textcolor{JungleGreen}{\bf\emph{if if you call the any eight hundred number}} if \textcolor{JungleGreen}{\bf\emph{you}} you can call up any eight hundred number }}\\
& \bf{Baseline} &\small{{\textcolor{JungleGreen}{\bf\emph{if}} if \hspace{0.02cm}you \hspace{0.02cm}call \hspace{0.02cm}the \hspace{0.02cm}any \hspace{0.02cm}eight \hspace{0.02cm}hundred \hspace{0.02cm}number if \textcolor{JungleGreen}{\bf\emph{you}} you can call up any eight hundred number }}\\
& \bf{Self-trained} & {\small {\textcolor{JungleGreen}{\bf\emph{if if you call the any eight hundred number}}  if \textcolor{JungleGreen}{\bf\emph{you}} you can call up any eight hundred number }}\\ \midrule

\multirow{3}{*}{\bf 2}  & \bf{Gold} & \small {{ \textcolor{JungleGreen}{\bf\emph{she was going to get picked up}} she was going to pick him up because she only $\cdots$}}\\
& \bf{Baseline} & {\small{she was going \hspace{0.02cm}to \hspace{0.02cm}get \hspace{0.02cm}picked \hspace{0.02cm}up she was going to pick him up because she only $\cdots$}}\\
& \bf{Self-trained} & \small {{\textcolor{JungleGreen}{\bf\emph{she was going to get picked up}} she was going to pick him up because she only $\cdots$}}\\ \midrule

\multirow{3}{*}{\bf 3} & \bf{Gold} & \small{{It goes back to you know \textcolor{JungleGreen}{\bf\emph{what right}} what can society impose on people}}\\
& \bf{Baseline} & \small{{It goes back to you know what right what can society impose on people}}\\
& \bf{Self-trained} & \small{{It goes back to you know \textcolor{JungleGreen}{\bf\emph{what}} right what can society impose on people}}\\ \midrule

\multirow{3}{*}{\bf 4} & \bf{Gold} & \small{and the money they do have \textcolor{JungleGreen}{\bf\emph{they're not}} they do not use it wisely}\\
& \bf{Baseline}  &\small{and the money they do have \hspace{0.02cm}they're \hspace{0.02cm}not \hspace{0.02cm}they do not use it wisely}\\
& \bf{Self-trained} & {\small {and the money they do have \textcolor{JungleGreen}{\bf\emph{they're not}} they do not use it wisely}}\\ \midrule

\multirow{3}{*}{\bf 5} & \bf{Gold} & \small {{For two years we didn't \textcolor{JungleGreen}{\bf\emph{and we}} which was  \textcolor{JungleGreen}{\bf\emph{a} }kind of stupid}}\\
& \bf{Baseline} & \small {{For two years we didn't and we which was a kind of stupid}}\\  
& \bf{Self-trained} & \small {{For two years we didn't \textcolor{JungleGreen}{\bf\emph{and we}} which was  a kind of stupid}}\\ \midrule

\multirow{3}{*}{\bf 6} & \bf{Gold} & \small {{\textcolor{JungleGreen}{\bf\emph{We}} we couldn't survive \textcolor{JungleGreen}{\bf\emph{in a in a juror}} in a trial system without a jury} }\\
& \bf{Baseline} &\small { {\textcolor{JungleGreen}{\bf\emph{We}} we couldn't survive \textcolor{JungleGreen}{\bf\emph{in a}} \hspace{0.02cm}in \hspace{0.02cm}a \hspace{0.02cm}juror \hspace{0.02cm}in a trial system without a jury}}  \\ 
& \bf{Self-trained} & \small {{\textcolor{JungleGreen}{\bf\emph{We}} we couldn't survive \textcolor{JungleGreen}{\bf\emph{in a in a juror}} in a trial system without a jury}  } \\  \midrule

\multirow{3}{*}{\bf 7} & \bf{Gold} &\small { {$\cdots$ I think it's like\hspace{0.035cm} ninety-nine \hspace{0.035cm}point \hspace{0.035cm}ninety-nine think it is}} \\
& \bf{Baseline} &\small {{$\cdots$ I think it's like \textcolor{Apricot}{\bf\ul{ninety-nine point}} ninety-nine think it is}}\\
& \bf{Self-trained} & \small {{$\cdots$ I think it's like \hspace{0.035cm}ninety-nine \hspace{0.035cm}point \hspace{0.035cm}ninety-nine think it is} }\\ \midrule

\multirow{3}{*}{\bf 8} & \bf{Gold} & \small {{Do you think for a big \hspace{0.02cm}or \hspace{0.02cm}a little place}}\\
& \bf{Baseline} & \small {{Do you think for \textcolor{Apricot}{\bf\ul{a big or}} a little place}}\\
& \bf{Self-trained} &\small {{Do you think for a big \hspace{0.02cm}or \hspace{0.02cm}a little place}}\\ \midrule

\end{tabular}
\end{center}
\caption{Some examples from the Switchboard dev set and corresponding EDITED disfluency labels given by the baseline and the best self-trained parser, as well as the gold (i.e. correct) labels. Green (and italic) words indicate correctly labeled disfluent words and orange (and underlined) words represent fluent words which are incorrectly labeled as disfluencies.}\label{tab:10} 
\end{table*}

\begin{figure*}[t]
\centering
\begin{subfigure}{.5\textwidth}
  \centering
  \includegraphics[width=.98\linewidth]{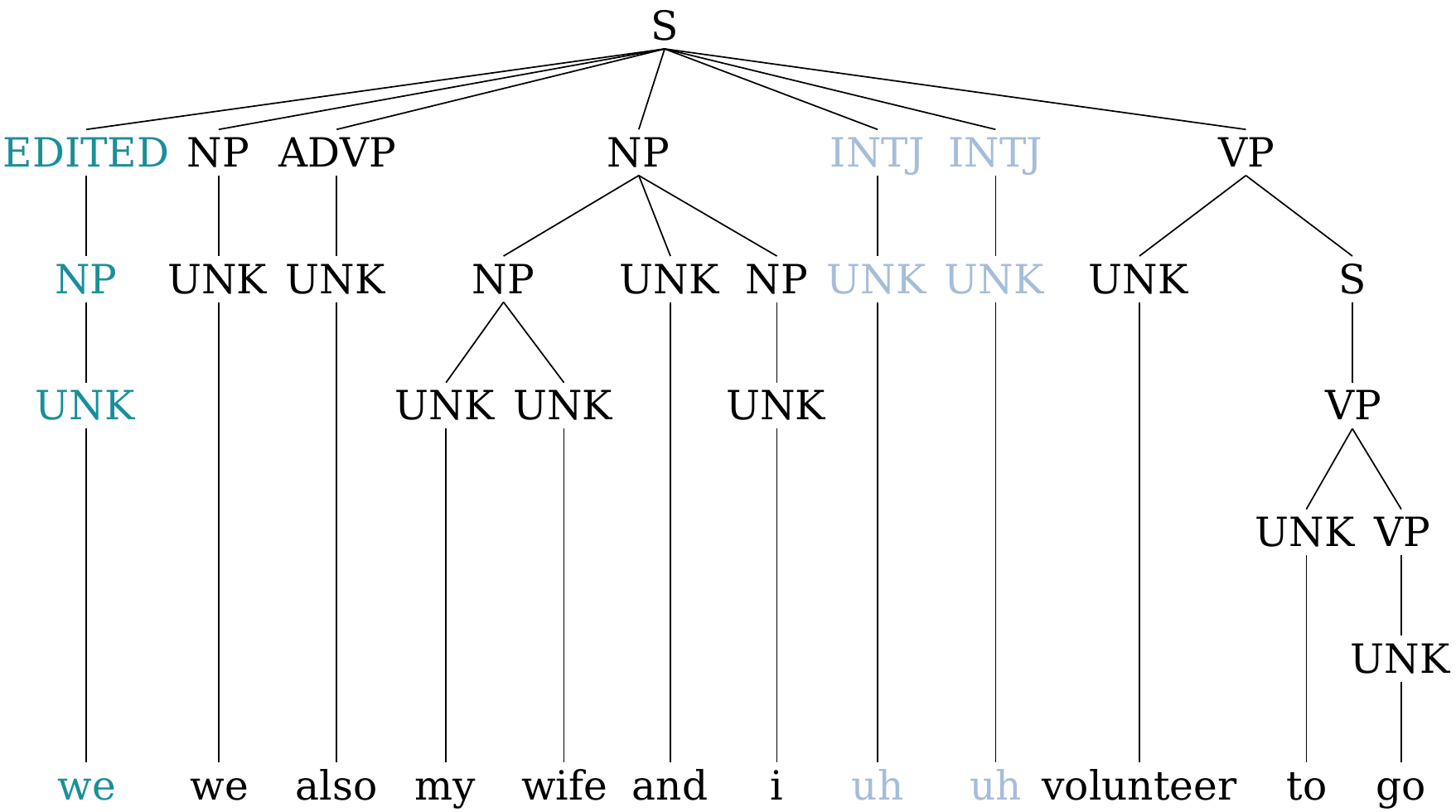}
  \caption{Baseline}
  \label{fig:sub1}
\end{subfigure}%
\begin{subfigure}{.5\textwidth}
  \centering
  \includegraphics[width=.98\linewidth]{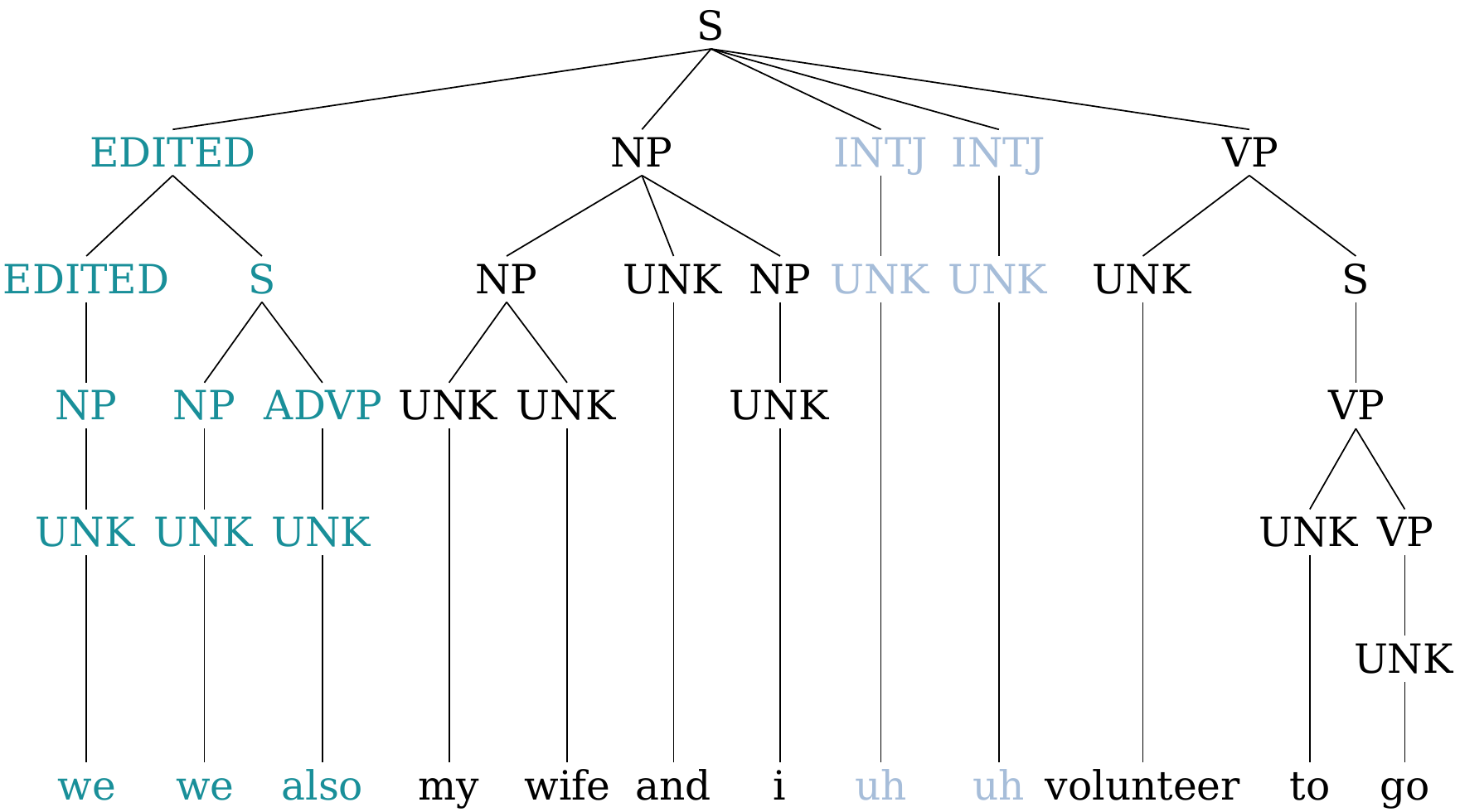}
  \caption{Self-trained}
  \label{fig:sub2}
\end{subfigure}\vspace{0.3cm}

\caption{A sentence from the Switchboard dev set parsed by the baseline model (left) and by the self-trained model (right). The parse tree obtained by the self-trained model is the same as the gold parse tree. }
\label{fig:test}
\end{figure*}

\subsection{Does ensembling parsers improve disfluency detection?}
We investigate the impact of ensembling on the performance of the self-attentive parser, where we combine parsers by averaging their span label scores as follows: 

\begin{equation} \label{eq:02}
s_{ensemble}(i,j,l)=\frac{1}{4}\sum_{n=1}^{4}s_{n}(i,j,l) 
\end{equation}

We tried different ensembling of parsers and the best result was achieved when we trained the baseline parser four times using four BERT word representations, namely BERT$_{\text{BASE}~ \text{[cased}\mid\text{uncased]}}$ and BERT$_{\text{LARGE}~\text{[cased}\mid\text{uncased]}}$, and combined the results at inference time~\citep{kita:19}. The ensembled models not only reflect variations of different pre-trained representations but also the randomness in initialization of the models. As shown in Table~\ref{tab:30}, ensembling and self-training both improve the performance of the baseline single model on parsing and detecting EDITED disfluency nodes. Self-training is more effective than ensembling, especially for EDITED node detection. The best results are reported for ensembling the best of the self-trained parsers for each of different BERT models from the 5 random restarts\footnote{We also tried ensembling all 20 versions of the self-trained parser initialized with different random seeds. The results were 0.1\% worse than the ensemble of the four best self-trained parsers.}.

\begin{table}[H]
\begin{center}
\begin{tabular}{p{3.6cm}p{0.59cm}p{0.75cm}p{1.1cm}}
\midrule \bf Model &  \small  $\bf F(S_E)$  & \small $\bf F(W_E)$  & \small $\bf F(W_{EIP})$ \\ \midrule
Baseline (single) & ~$89.2$ & ~~~$90.9$ & ~~~$95.3$\\ 
Baseline (ensemble) & ~$90.3$& ~~~$91.1$ & ~~~$95.6$\\ 
Self-trained (single) & ~$90.7$ & ~~~$92.4$ & ~~~$96.0$\\ 
Self-trained (ensemble)   & ~$90.9$ & ~~~$92.8$ & ~~~$96.4$\\ \midrule
\end{tabular}
\end{center}
\caption{Parse f-score for EDITED node \emph{F($\text{S}_\text{E}$)}, EDITED word f-score \emph{F(W$_{E}$)} and EDITED, INTJ and PRN word f-score \emph{F(W$_{EIP}$)} for different models on the Switchboard dev set. ``single''= single parser and ``ensemble''= ensemble of 4 parsers.}\label{tab:30} 
\end{table}




\section{Results}
We compare the performance of our best model with previous work on the Switchboard test set. As demonstrated in Table~\ref{tab:05}, our model outperforms prior work in parsing. The parsing result for our model is higher than~\citet{trang:18} which utilizes prosodic cues, as well as text based features. 

\vspace{-0.143cm}
\begin{table}[H]
\begin{center}
\begin{tabular}{lccc}
\midrule\bf Parsing (S) & \bf P & \bf R & \bf F  \\ \midrule 
 \small\citet{trang:18} &  $~-$ & $~-$ &$87.9$ \\ 
 \small\citet{trang:18}$^\ast$  &  $~-$ & $~-$ & $88.5$ \\
 \small\citet{jam:19}  & $92.4$ & $92.9$ & $92.7$ \\ 
  \small\citet{tran:19} & $~-$ & $~-$ & $92.8$    \\ 
  \small\citet{tran:19}$^\ast$  & $~-$ & $~-$ & $93.0$   \\  
 \small{This work (single model)}&  $93.2$ & $93.8$ & $93.5$\\
  \small{This work (ensemble of 4)} &  $93.6$ & $94.2$ & $93.9$  \\ \midrule 
\end{tabular}
\end{center}
\caption{\label{tab:05} Parse precision \emph{P}, recall \emph{R} and f-score \emph{F} for all constituent spans on the Switchboard test set. $^\ast$Text+prosody model. P=P(S), R=R(S) and F=F(S). 
}
\end{table}

We compare the performance of the self-attentive parser with state-of-the-art disfluency detection models. As shown in Table~\ref{tab:06}, our model has the best f-score. We also compare our model with prior work that reported EDITED, INTJ and PRN word f-score for disfluency detection and find that our model has the best performance (see Table~\ref{tab:07}). Compared to~\citet{wang:18} which uses GANs to leverage additional unlabelled data and~\citet{bach:19} which leverages synthetic data, our model significantly improves the recall. This demonstrates that standard techniques such as self-training and ensembling are as good or better than these specialized, complex approaches. 

\begin{table}[H]
\begin{center}
\begin{tabular}{lccc}
\midrule  \bf Disfluency (E) & \bf P & \bf R & \bf F \\ \midrule 
\small\citet{trang:18}   &  $~-$ &  $~-$ &   $76.7$  \\ 
\small\citet{trang:18}$^\ast$  & \small$~-$ &  $~-$ &  $77.5$ \\ 
\small\citet{jam:18}\tablefootnote{\url{https://github.com/pariajm/deep-disfluency-detector}} &   $89.5$ &  $80.0$ &   $84.5$  \\ 
\small\citet{zay:16} &   $91.8$ &  $80.6$ &  $85.9$  \\   
\footnotesize\citet{jam:17} &  $~-$ &  $~-$ & $86.8$ \\  
\small\citet{wang:16} &  $91.6$ &  $82.3$ &  $86.7$ \\   
\small\citet{jam:19} &  $81.7$ &  $92.8$ & $86.9$  \\
\small\citet{wang:17} &  $91.1$ &  $84.1$ &  $87.5$  \\   
\small\citet{dong:19}~ &  $94.5$ & $84.1$ & $89.0$ \\ 
\small{This work (single model)} & $86.7$ & $91.9$ & $89.2$\\ 
\small{This work (ensemble of 4)}& $87.5$ & $93.8$ & $90.6$
   \\
\midrule

\end{tabular}
\end{center}
\caption{\label{tab:06} EDITED word precision \emph{P}, recall \emph{R} and f-score \emph{F} on the Switchboard test set. $^\ast$Text+prosody model. P={{P(W$_\text{E}$)}}, R={{R(W$_\text{E}$)}} and F={{F(W$_\text{E}$)}}. 
}
\end{table}

\begin{table}[h]
\begin{center}
\begin{tabular}{lccc}
\midrule  \bf Disfluency (EIP) & \bf P & \bf R & \bf F \\ \midrule
 \small\citet{wang:18}   &  $92.1$ &  $90.2$ &   $91.1$  \\ 
\small\citet{bach:19}  & $94.7$ & $89.8$ &  $92.2$ \\  
\small{This work (single model)} & $92.2$ & $96.6$ & $94.3$  \\ 
\small{This work (ensemble of 4) }& $92.5$ & $97.2$ & $94.8$ 
  \\ 
\midrule

\end{tabular}
\end{center}
\caption{\label{tab:07} EDITED, INTJ and PRN (EIP) word precision \emph{P}, recall \emph{R} and f-score \emph{F} on the Switchboard test set. P={{P(W$_\text{EIP}$)}}, R={{R(W$_\text{EIP}$)}} and F={{F(W$_\text{EIP}$)}}. 
}
\end{table}

\subsection{Qualitative Results}
We conduct a qualitative analysis on the Switchboard dev set to characterize the disfluencies that the baseline model cannot detect but the self-trained one can. We provide representative examples  in Table~\ref{tab:10}. In general, the self-trained model is better at detecting long complex corrections (\# 1-4), restarts (\# 5) and stutter-like repetitions (\# 6). It also does a better job of discriminating fluent repetitions and fluent parallel structures from repetition and correction types of disfluency (\# 7 and 8). Figure~\ref{fig:test} depicts a sentence parsed by the baseline and the self-trained self-attentive parser, where the self-trained model correctly predicts all disfluency EDITED nodes. As explained in Section~\ref{sec:mo}, we do not use an external POS tagger, so POS tags are not available when parsing from raw text. That's why all preterminal labels in Figure~\ref{fig:test} are shown by a dummy token i.e. UNK.



\section{Conclusion}
We introduced a new state-of-the-art for joint disfluency detection and constituency parsing of transcribed speech. We showed that self-training and ensembling are effective methods for improving disfluency detection. A qualitative analysis of the results also indicated that self-training is helpful for detecting complicated types of disfluencies, including corrections and restarts. 
In future work, we intend to explore the idea of self-training for parsing written texts. We also aim at integrating syntactic parsing and self-training more closely with automatic speech recognition. The first step is to develop parsing models that parse ASR output, rather than speech transcripts. 



\section*{Acknowledgments}
We would like to thank the anonymous reviewers for their insightful comments and suggestions. This research was supported by a Google award through the Natural Language Understanding Focused Program, by a CSIRO's DATA61 Top-up Scholarship, and under the Australian Research Councils Discovery Projects funding scheme (project number DP160102156). 
\bibliography{anthology,acl2020}
\bibliographystyle{acl_natbib}




\end{document}